\documentclass[11pt,a4paper]{article}
\usepackage[hyperref]{acl2021}
\usepackage{times}
\usepackage{latexsym}

\usepackage{microtype}

\usepackage{multirow}
\usepackage{CJKutf8}
\usepackage{amsmath}
\usepackage{xspace}
\usepackage{graphicx}
\usepackage[switch]{lineno}
\usepackage{enumitem}
\usepackage{paralist}
\usepackage{tabulary}
\usepackage{booktabs}
\usepackage{color}

\usepackage{caption}
\usepackage{subcaption}

\aclfinalcopy % Uncomment this line for the final submission
\title{The Volctrans Neural Speech Translation System for IWSLT 2021}

\author{Chengqi Zhao\quad Zhicheng Liu\quad Jian Tong\quad Tao Wang\quad Mingxuan Wang\\
    \textbf{Rong Ye\quad Qianqian Dong\quad Jun Cao\quad Lei Li} \\
  ByteDance AI Lab \\
  \texttt{\{zhaochengqi.d,liuzhicheng.lzc,tongjian,wangtao.960826 }\\
  \texttt{wangmingxuan.89,yerong,dongqianqian} \\
  \texttt{caojun.sh,lileilab\}@bytedance.com} \\}

\date{}

\begin{document}
\maketitle
\begin{abstract}
This paper describes the systems submitted to IWSLT 2021 by the Volctrans team.
We participate in the offline speech translation and text-to-text simultaneous translation tracks. 
For offline speech translation, our best end-to-end model achieves 7.9 BLEU improvements over the benchmark on the MuST-C test set and is even approaching the results of a strong cascade solution. 
For text-to-text simultaneous translation, we explore the best practice to optimize the \texttt{wait-k} model. 
%Incremental encoders and better decoding strategies are applied to improve the adaptability of models to streaming input. 
As a result, our final submitted systems exceed the benchmark at around 7 BLEU on the same latency regime.
We release our code and model to facilitate both future research works and industrial applications\footnote{Code and models are available at \url{https://github.com/bytedance/neurst/tree/master/examples/iwslt21}}.
\end{abstract}

\section{Introduction}
\label{sec:intro}
This paper describes the neural speech translation systems submitted to IWSLT 2021 by the Volctrans team (also known as the team from ByteDance AI Lab), including cascade and end-to-end speech translation (ST) systems for the offline ST track and a simultaneous neural machine translation (NMT) system.  We aim at finding the best practice for these two tracks. 

For offline ST,  the cascaded system often outperforms the fully end-to-end approach. 
Recent studies on the fully end-to-end approaches obtain promising results and attract a lot of interest. 
%However, the most critical problem for end-to-end solutions comes from the lack of supervised training data.  Several attempts have been made to address the data scarcity problem, such as pre-training~\cite{Bansal2019pretrain,stoian2020pretrain,wang2020curriculum,alinejad2020effectively}, data augmentation~\cite{jia2019weak,pino2020self}, multi-task learning~\cite{Weiss2017e2e,Berard2018e2e,tang2020mtl,ye2021progressive} and knowledge distillation~\cite{Liu2019kd,Gaido2020kd,dong2021ted}. 
Last year's results have shown that an end-to-end model achieves an even better performance~\cite{ansari2020} compared with the cascaded competitors. However, they introduce  pre-training ~\cite{Bansal2019pretrain,stoian2020pretrain,wang2020curriculum,alinejad2020effectively} and data augmentation techniques~\cite{jia2019weak,pino2020self} to end-to-end models, while the cascaded is not that strong enough.  
Hence, in this paper, we would like to optimize the speech translation model in two aspects.
% Hence, it is still not the time to conclude that end-to-end models have overwhelmed cascade models.  
% Different from previous studies,  we would like to investigate this problem in two aspects.
First, we are devoted to building a strong cascade competitor and learns the best practice from WMT evaluation campaigns~\cite{li-etal-2019-niutrans,wu2020volctrans}, such as back translation~\cite{sennrich2016bt} and ensemble. 
Second, we explore various self-supervised learning methods and introduce as much semi-supervised data as possible towards finding the best practice of training end-to-end ST models.  
In our settings, ASR data, MT data, and monolingual text data are all considered in a progressively training framework. 
The results are very promising, and the final performance on the MuST-C test set surpasses the end-to-end baseline by 7.9 BLUE scores, while it is still lagging behind our cascade model by 1.5 BLUE scores. 
It is not surprising since some well-optimized methods for MT can not be easily used on ST, such as back translation. 
However, our experience shows that the external data can effectively close the gap between end-to-end models and cascade models.

%In this paper, we would like to investigate the performance of the end-to-end solutions with massive speech-translation pairs by data augmentation. And we also explore the potential of the multi-task learning framework, which can involve a large number of additional ASR and MT data. Meanwhile, we learn from the experience of WMT evaluation campaigns~\cite{li-etal-2019-niutrans,wu2020volctrans} to build a much stronger MT model for our cascade system. By comparing these two systems, we want to see whether the end-to-end approaches can become the dominant technology in ST and have been ready for industrial applications.

In parallel,  we also participate in the simultaneous NMT track, which translates in real-time. Our system is based on an efficient \texttt{wait-k} model~\cite{elbayad2020efficient}. 
We investigate large-scale knowledge distillation~\cite{Kim2016,Freitag2017} and back translation methods. 
Specially, we develop a \texttt{multi-path} training strategy, which enables a unified model  serving different \texttt{wait-k} paths. 
Our target is to obtain the best translation quality at different latency levels.

The remaining part of the paper proceeds as follows. Section~\ref{sec:cascade} and section~\ref{sec:e2e} describe our cascade and end-to-end systems respectively. Section~\ref{sec:simul} presents the implementation of simultaneous NMT models. Each section starts from the training sources and how we synthesize large-scale data. And then, we give details about the model structure and techniques for training and inference. We conduct experiments using only the provided datasets by IWSLT 2021, and results are shown in Section~\ref{sec:exp}.

\section{Cascaded Speech Translation}
\label{sec:cascade}
\subsection{Automatic Speech Recognition} \label{sec:asr}
The ASR model is transformer-like and trained on paired speech and transcript data 

\paragraph{Datasets and Preprocessing}

We divide the allowed ASR datasets into two parts: clean and noisy and consider MuST-C\footnote{In this paper, MuST-C denotes the newly released English-German ST dataset~(v2) by IWSLT 2021.}, LibriSpeech~\cite{Panayotov2015}, and Mozilla Common Voice as the clean datasets, and use them for training an ASR system to filter the noisy part, i.e., \textit{iwslt-corpus}\footnote{The training corpus for IWSLT evaluation campaign over the last years.} and TED-LIUM 3~\cite{Hernandez2018}. We remove the training samples where the word error rate~(WER) score between the ASR output and English transcript exceeds 75\%. The statistics of the ASR datasets are shown in Table~\ref{tb:asr_statistics}.

\begin{table}[!t]\small%\setlength{\tabcolsep}{4pt}
  \centering
  \begin{tabular}{lcc}
    \toprule
    \textbf{Dataset} & \textbf{\#samples} & \textbf{\#hours}  \\
    \midrule
    MuST-C & 250,942 & 450 \\
    LibriSpeech & 281,241 & 961 \\
    Common Voice & 562,517 & 899 \\
    \textit{iwslt-corpus} &  157,909 & 231 \\
    TED-LIUM 3 & 111,600 & 165 \\
    \bottomrule
  \end{tabular}
  \caption{The statistics of audio datasets to train the ASR model. The \textit{iwslt-corpus} and TED-LIUM 3 are filtered by an ASR model trained on MuST-C, LibriSpeech and Common Voice.}
  \label{tb:asr_statistics}
\end{table}

For model training, we extract 80-channel log Mel-filterbank coefficients with windows of 25ms and steps of 10ms on the audio input. 
The transcripts are lowercased and we remove all punctuation marks. Then, we apply Moses tokenizer\footnote{\url{https://github.com/moses-smt/mosesdecoder/blob/master/scripts/tokenizer/tokenizer.perl}} and byte pair encoding (BPE)~\cite{sennrich2016}\footnote{\url{https://github.com/rsennrich/subword-nmt}} to the transcripts with 8,000 merge operations.

\begin{figure}[t] 
	\centering
	\includegraphics[angle=0,width=0.48\textwidth]{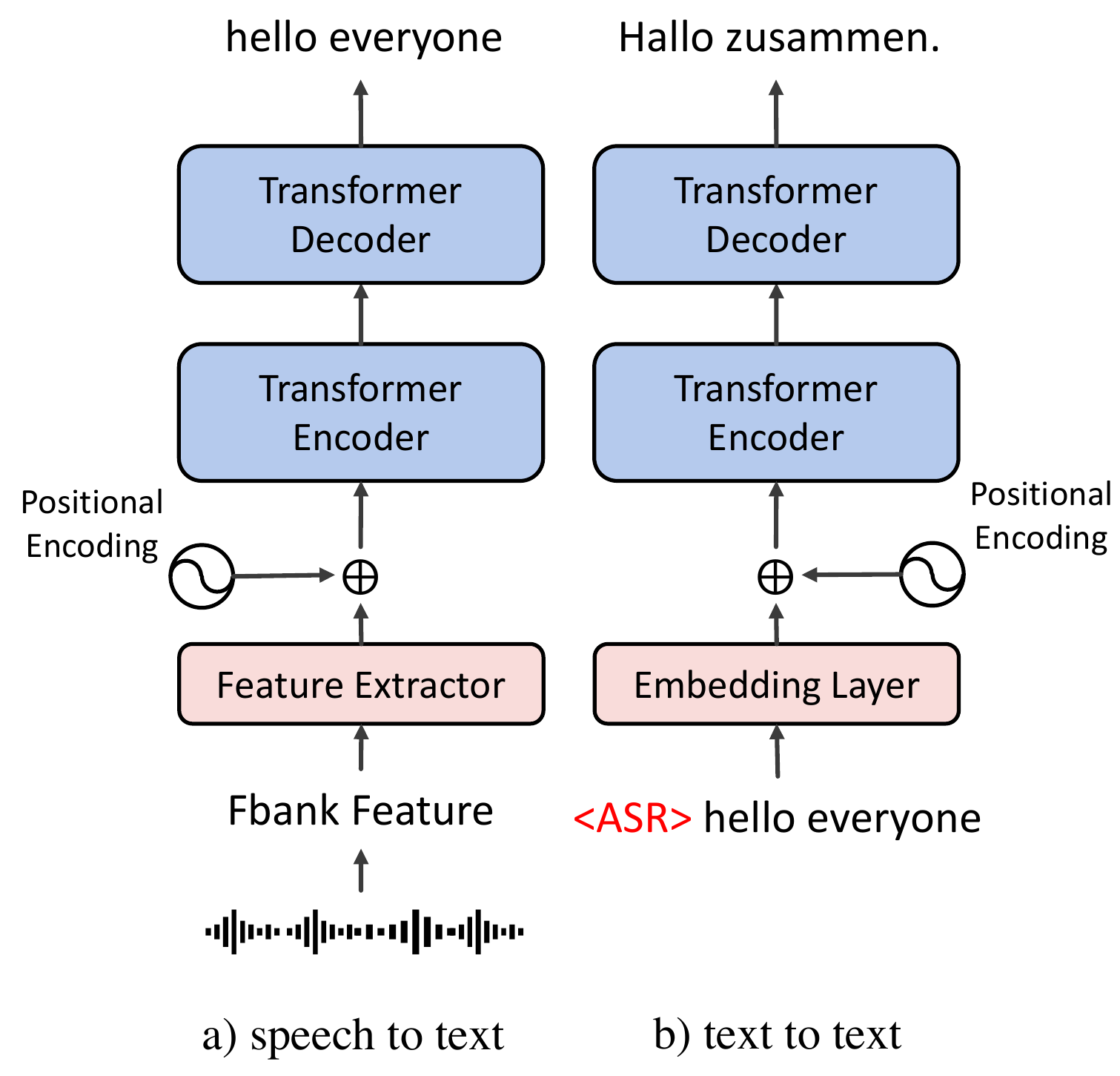}
	\caption{Overview of the cascaded speech translation model.}
	\label{fig:cascade_model}
\end{figure}

\paragraph{End-to-End ASR Model}
We refer to the recent progress of transformer-based ASR~\cite{Dong2018,Karita2019} and implement the speech transformer model, as illustrated in Figure~\ref{fig:cascade_model} a). 
The feature extractor consists of two-layer CNN with 256 channels, $3\times 3$ kernel, and stride size 2, each of which is followed by a layer normalization and ReLU activation. 
The major architecture is the same as the transformer model, including 12 layers for the encoder and 6 layers for the decoder. The model width is 768, and the hidden size of the feed-forward layer is 3,072. The attention head is set to 12 for both self-attention and cross-attention. 
To train the model, we use Adam optimizer~\cite{kingma2014adam} and set the warmup steps to 25,000. Empirically, we scale up the learning rate by 5.0 to accelerate the convergence. The ASR model is trained on 8 NVIDIA Tesla V100 GPUs with 320,000 frames per batch. 
And we truncate the audio frames to 3,000 and remove training samples whose transcript length exceeds 120 for GPU memory efficiency.
To further improve the performance, we apply SpecAugment technique~\cite{Daniel2019} with frequency masking ($mF=2, F=27$) and time masking ($mT=2, T=70, p=0.2$).

\subsection{Neural Machine Translation} \label{sec:cas_nmt}
All MT models are based on transformer \cite{vaswani2017}. We employ data augmentation and model ensemble techniques to improve the final performance.

\paragraph{Datasets and Preprocessing}
We utilize English-German~(EN-DE) parallel sentences from WMT 2020\footnote{\url{http://www.statmt.org/wmt20/translation-task.html}, including Common Crawl, Europarl v10, News Commentary v15, and ParaCrawl v5.1 }, OpenSubtitles 2018\footnote{\url{https://opus.nlpl.eu/OpenSubtitles2018.php}}, MuST-C and \textit{iwslt-corpus} for training.
We filter the parallel corpora following the rules listed in \citet{li-etal-2019-niutrans}, with a much stricter constrain on word alignment. 
Additionally, we randomly select 10\% sentences separately from both sides of the original WMT and OpenSubtitles corpus for data augmentation~(see below), along with the transcripts in ASR datasets described in sec~\ref{sec:asr}. 

As for text preprocessing, we apply Moses tokenizer and BPE with 32,000 merge operations on each side.

\paragraph{Tagged Back-Translation}
Back-translation \cite{sennrich2016bt} is an effective way to improve the translation quality by leveraging a large amount of monolingual data and has been widely used in WMT evaluation campaigns. 
In our setting, we add a ``$<$BT$>$" tag to the source side of back-translated data to prevent overfitting on the synthetic data, which is also known as tagged back-translation~\cite{Caswell2019,Marie2020}.

\paragraph{Knowledge Distillation}
Sequence-level knowledge distillation \cite{Kim2016,Freitag2017} is another useful technique to improve performance. In this way, we enlarge the training data by translating English sentences to German using a good teacher model.

\paragraph{ASR Output Adaptation}
Traditionally, the output of ASR systems is lowercased with no punctuation marks, while the MT systems receive natural texts. 
In our system, we attempt to make the MT systems robust to these irregular texts. 
A simple way to do so is to apply the same rules on the source side of the MT training set. However, empirical study shows it causes performance degradation. 
Inspired by the tagged back-translation method, we enhance the regular MT models with transcripts from both ASR systems and the ASR datasets, as illustrated in Figure~\ref{fig:cascade_model} b). An extra tag ``$<$ASR$>$" indicates the irregular input.
Note that the basic idea to bridge the gap between the ASR output and the MT input involves additional sub-systems, like case and punctuation restoration. In our cascade system, we prefer to use fewer sub-systems, and the detailed comparison would be our future work.

\begin{table*}[!t]\small%\setlength{\tabcolsep}{4pt}
  \centering
  \begin{tabular}{lccccccc}
    \toprule
    \multirow{2}{*}{\textbf{Dataset}} & \multirow{2}{*}{\textbf{Size}} & \multirow{2}{*}{\textbf{MT\#1}} & \multicolumn{2}{c}{\textbf{MT\#2}} & \multirow{2}{*}{\textbf{MT\#3}} & \multirow{2}{*}{\textbf{MT\#4}} & \multirow{2}{*}{\textbf{MT\#5}} \\
    \cline{4-5}
     & & & \textbf{pretrain} & \textbf{fine-tune} & & &  \\
    \midrule
    WMT 2020 & 13.7M & P & P & / & P & / & P \\
    OpenSubtitles 2018 & 10.7M & P & P & / & P & P & / \\
    MuST-C & 0.25M & P & P/BT/SR & P/BT/SR & P/SR/KD & P/BT/SR & P/BT/SR \\
    \textit{iwslt-corpus} & 0.16M & / & P/BT/SR & P/BT/SR & P/SR/KD & P/SR & P/BT/SR \\
    TED-LIUM 3 (EN) & 0.11M & / & / & / & KD & / & / \\
    Common Voice (EN) & 0.56M & / & / & / & KD & / & / \\
    extra monolingual (EN/DE) & 6.77M &  / & / & BT & KD & BT & BT \\
    \bottomrule
  \end{tabular}
  \caption{The statistics of MT datasets after data filtering and the detailed combination modes of datasets for difference MT models~(MT\#1-5). The MT\#1 setting is used for training both DE$\rightarrow$EN and EN$\rightarrow$DE directions. ``P" denotes the parallel corpus. ``BT" is the back-translated data using MT\#1 (DE$\rightarrow$EN). ``SR" indicates the irregular data from both ASR datasets and the ASR model. ``KD" is the synthetic data generated by MT\#2.}
  \label{tb:mt_training_sets}
\end{table*}

\paragraph{Data Combination and Sampling Strategy}
We train transformer models with different combinations of data sets because increasing the model's diversity can benefit the model ensemble. The detailed setups are listed in Table~\ref{tb:mt_training_sets}. We over-sample the in-domain datasets (i.e., MuST-C/\textit{iwslt-corpus}-related portions) to improve the in-domain performance. Specifically, to control the ratio of samples from different data sources, we sample a fixed number of sentences being proportional to $(\frac{N_s}{\sum_s{N_s}})^{\frac{1}{T}}$, where $N_s$ is the number of sentences from data source $s$, and sampling temperature $T$ is set to 5. 
Note that the MT\#1 is trained on lowercased source texts without punctuation marks, while MT\#2-5 use the tagged transcripts.

\paragraph{Model Setups}
We follow the transformer big setting, except that
\begin{compactitem}
\item we deepen the encoder layers to 16.
\item the dropout rate is set 0.15.
\item the model width is changed to 768, the hidden size of the feed-forward layer is 3,072, and the attention head is 12 for MT\#5 only.
\end{compactitem}
We use Adam optimizer with the same schedule algorithm as \citet{vaswani2017}. All models are trained with a global batch size of 65,536. 

\subsection{Inference}
We average the latest 10 checkpoints of a single training process for all the above experiments. And during inference, the ``$<$ASR$>$" tag is added to the front of the ASR output. The beamwidth is set to 10 for both ASR and MT tasks.

\section{End-to-End Speech Translation}
\label{sec:e2e}
Recent studies show that the fully end-to-end solution achieves promising performance when compared with the cascaded models~\cite{ansari2020}. This section will introduce how we build our end-to-end models for the offline ST task.

\subsection{Training Data}

\begin{table}[!t]\small%\setlength{\tabcolsep}{4pt}
  \centering
  \begin{tabular}{lrr}
    \toprule
    \textbf{Dataset} & \textbf{\#samples} & \textbf{\#hours}  \\
    \midrule
    MuST-C & 1,198,056 & 2,186 \\
    \textit{iwslt-corpus} & 746,714 & 1,112 \\
    LibriSpeech & 1,117,394 & 3,833  \\
    Common Voice & 2,212,581  & 3,546  \\
    TED-LIUM 3 & 384,389 & 577 \\
    \bottomrule
  \end{tabular}
  \caption{The size of audio datasets with data augmentation to train the end-to-end ST model. }
  \label{tb:e2e_statistics}
\end{table}

The end-to-end model is trained on paired speech and translation data. We collect MuST-C and \textit{iwslt-corpus}~(after filtering described in section~\ref{sec:cascade}), with a total of only 681 hours transcribed and translated speech. To address the data scarcity problem, we explore the knowledge distillation technique to augment the data by leveraging ASR datasets and MT models, also known as pseudo labeling. In detail, we distill from four MT models: MT\#1, MT\#2, an ensemble of MT\#3-5, and MT\#3-R2L which is trained with the same setting as MT\#3 and generates the target translations in the right to left fashion. We filter the augmented samples with bad alignment scores as the same as data filtering in MT. The statistics of training data is shown in Table~\ref{tb:e2e_statistics}.

Moreover, two additional copies of the original and the augmented training data are created by modifying the speed to 110\% and 90\% of the initial rate, which makes a 3-fold training set.

\subsection{Speech Transformer for End-to-End ST}
As a baseline system, the model architecture and training configurations are the same as the end-to-end ASR in our cascade system, except for the learning rate, which is scaled up by 3.0 for ST.
We initialize the feature extractor and encoder from the corresponding component of ASR.

We keep the cases and punctuation marks on the target side and apply Moses tokenizer and BPE to the translations with 32,000 merge operations.

\subsection{Progressive Multi-task Learning}
\begin{figure}[t] 
	\centering
	\includegraphics[angle=0,width=0.45\textwidth]{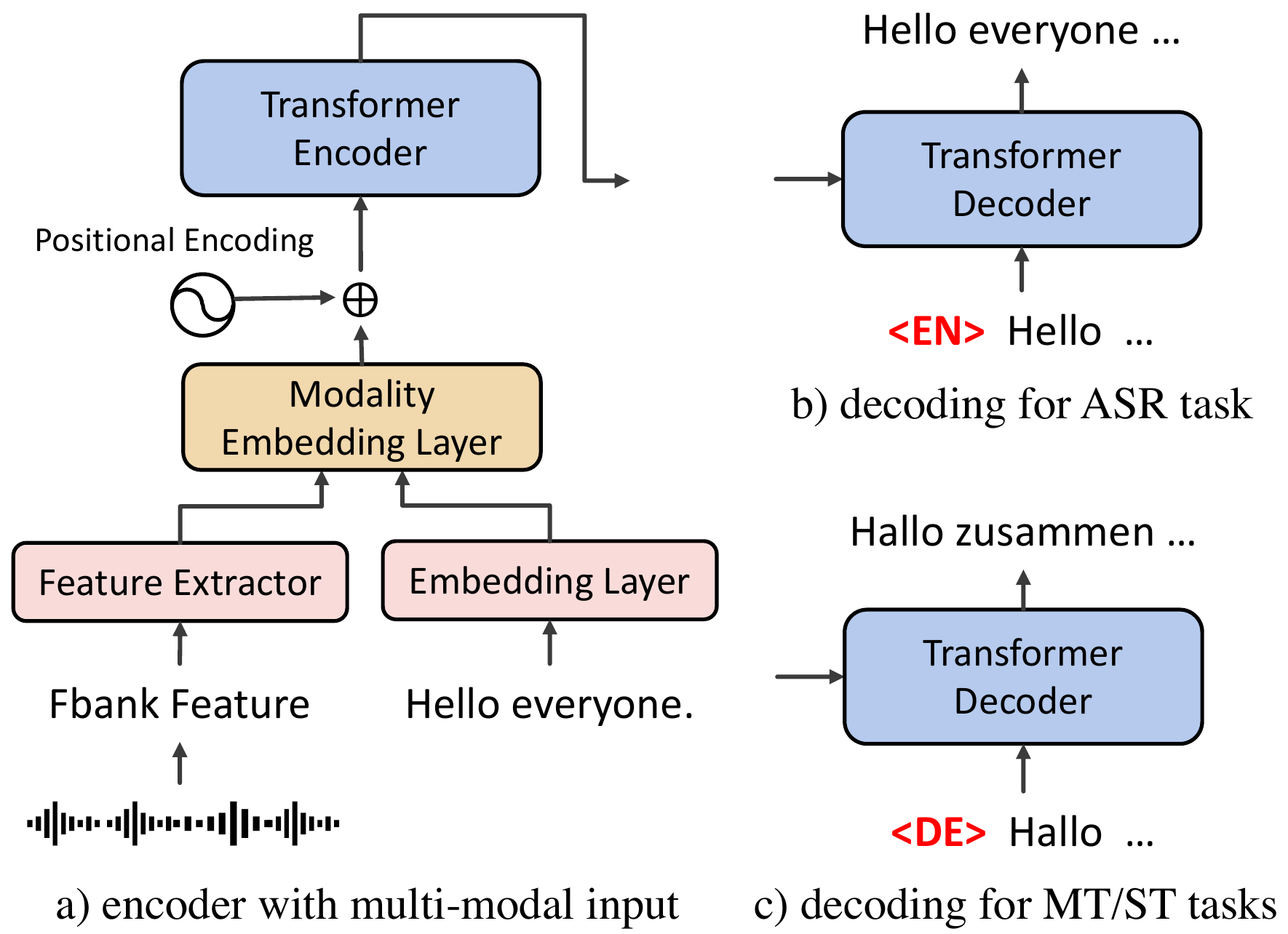}
	\caption{Overview of the end-to-end ST model with progressive multi-task learning. Note that the audio and text inputs are unnecessary to be aligned during training.}
	\label{fig:multi_model_task_st}
\end{figure}

Inspired by the multi-task learning framework for ST and the progressive training strategy \cite{tang2020mtl,ye2021progressive}, we introduce PMTL-ST, a progressive multi-task learning framework for speech translation, which can leverage additional ASR and MT data for training. As illustrated in Figure~\ref{fig:multi_model_task_st} a), the encoder accepts both audio and text inputs. Then we add a modality embedding to the representation to indicate audio input or text before passing to the shared transformer encoder. For decoding, we involve ``$<$EN$>$" and ``$<$DE$>$" tokens to make the decoder compatible with ASR and translation~(MT/ST) tasks, as shown in~\ref{fig:multi_model_task_st} b)/c).

For progressive training, we separately train an ASR model and an MT model via different branches in Figure~\ref{fig:multi_model_task_st}. Then, we initialize the feature extractor and the audio modality embedding from the ASR model, and the rest of the model parameters are initialized by the MT model. The final model is trained jointly with ASR, MT, and ST.

All other training configurations, such as batch size and learning rate, are the same as the corresponding single task described before. Additionally, for the PMTL-ST models, we jointly learn the sentencepiece\footnote{\url{https://github.com/google/sentencepiece}} model with 16,000 tokens on the mixture of English and German texts.

\subsection{Fbank2vec}
\begin{figure}[t] 
	\centering
	\includegraphics[angle=0,width=0.26\textwidth]{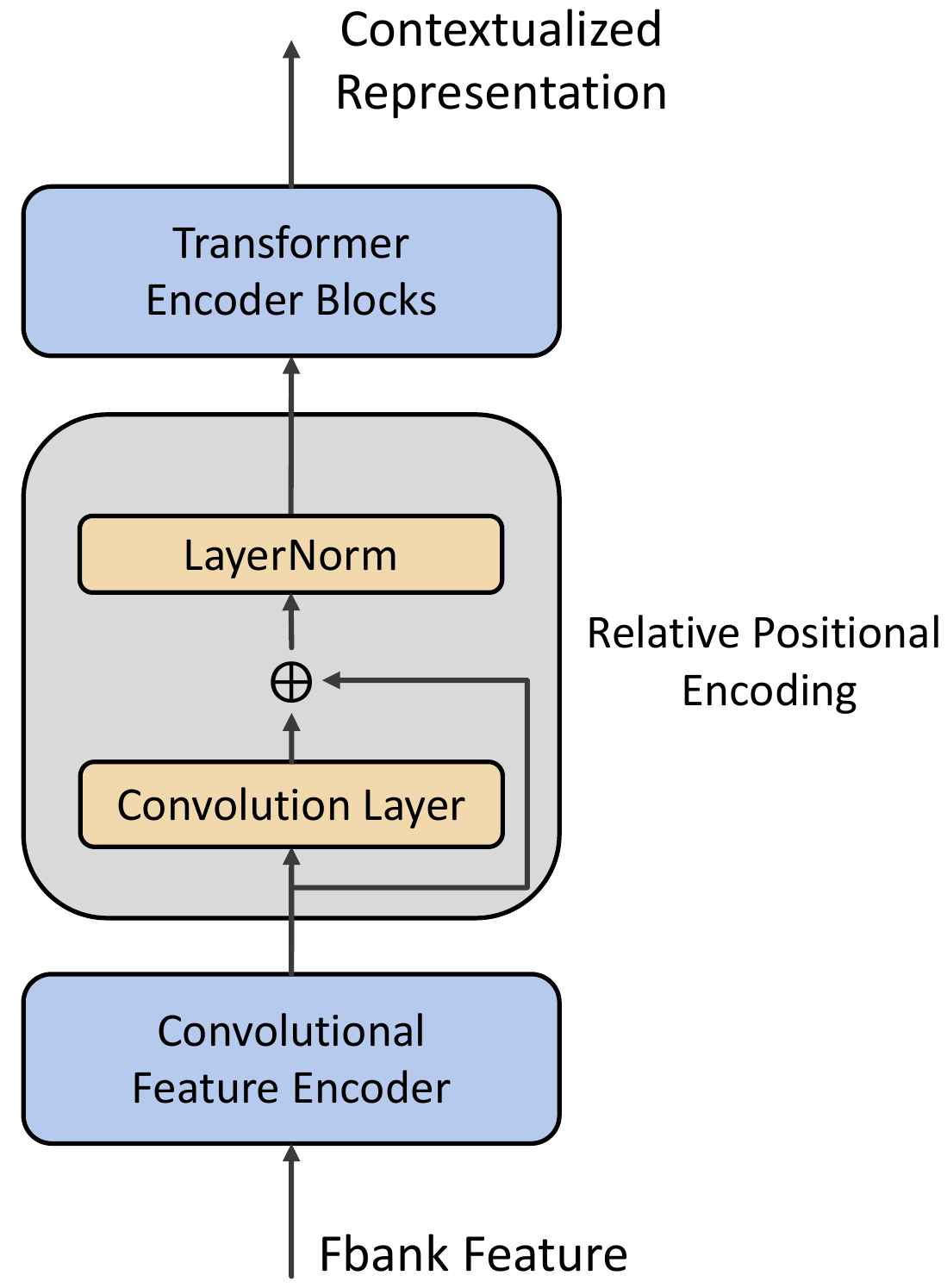}
	\caption{The proposed fbank2vec network for audio feature encoding.}
	\label{fig:fbank2vec}
\end{figure}

Inspired by the recent progress of speech representation learning, like wav2vec 2.0~\cite{Baevski2020}, we introduce a fbank2vec network to learn contextualized audio representations from log Mel-filterbank features, as shown in Figure~\ref{fig:fbank2vec}.

\paragraph{Convolutional Feature Encoder} 
The encoder consists of two blocks containing a convolution followed by layer normalization and a GELU activation~\cite{Hendrycks2016}. The convolution in each block has 512 channels with 3$\times$3 kernel and stride size 2.
\paragraph{Relative Positional Encoding} 
We use a group convolution layer to model the relative positional embeddings as \citet{Baevski2020} does. The kernel size is 128, and the number of groups is 16. 

\paragraph{Contextualized Encoder} The final contextualized audio representations are generated by several transformer encoder blocks. In our setting, we stack 6 layers of the post-norm transformer, and the inner activation function for the feed-forward layers is GELU. In turn, the number of shared encoder layers in Figure~\ref{fig:multi_model_task_st} is changed to 6.

We insert the fbank2vec network in the front of the feature extractor. The feature extractor further reduces the dimension of audio representations by one convolution layer with 5$\times$5 kernel and stride size 2. The number of channels keeps the same as the dimension of fbank2vec output. 

We experiment with two setups, fbank2vec-768 and fbank2vec-512.
The fbank2vec-768 means that 
\begin{compactitem}
\item the dimension of fbank2vec output is 768;
\item inner the contextualized encoder, the hidden size of feed-forward layers is 3,072, and the head of the self-attention layers is 12.
\end{compactitem}
For the fbank2vec-512, the numbers are 512, 2,048, and 8, respectively. 
Note that the fbank2vec module is pretrained by an ASR task and the overall model follows the progressive multi-task learning framework, so the configurations of word embeddings, the shared encoder and decoder vary accordingly.

\section{Simultaneous Translation}
\label{sec:simul}
This section describes our submissions to the text-to-text simultaneous speech translation track for English to German (EN2DE) and English to Japanese (EN2JA).
For versatility, we adopt identical methods for these two language pairs.

% This section is organized as follows:
% We first introduce data processing scheme in section \ref{simul:data}. Then, we describe the model architecture in section \ref{simul:model}. We detail our experimental setup and tricks applied during training and inferring in section \ref{simul:train-infer}. We finally report the results on two different language pairs in section \ref{simul:results}.

\subsection{Training Data}
\label{simul:data}
% total data; filter ; ftbt ; partition
The training data for EN$\rightarrow$DE is from MuST-C, OpenSubtitles 2018, and WMT 2020 datasets. And for EN$\rightarrow$JA, we use the parallel and monolingual data from the WMT 2020 news task. 
% For NMT training, we utilize the parallel data allowed for the IWSLT 2021.
% For English to German, we collect MuST-C, OpenSubtitles 2018, and WMT 2020 datasets.
% For English to Japanese, we use the parallel and monolingual data from the WMT 2020 news task. 

\paragraph{Data Preprocessing}
We follow the data filtering process proposed in WMT works~\cite{li-etal-2019-niutrans,wu2020volctrans}, including language detection, length ratio filtering, dictionary alignment, and so on. 
For pre-processing, we first apply MeCab\footnote{\url{https://github.com/taku910/mecab}} tokenizer to the Japanese sentences.
Then, words are segmented into subword units using sentencepiece toolkit for both language pairs. We jointly learn on the source and target side with a vocabulary of 10,000 tokens. 

\begin{table*}[!t]\small
  \centering
  \begin{tabular}{lccccccc}
    \toprule
    \textbf{Dataset} & \textbf{Size} & \textbf{MT\#0} & \textbf{MT\#1} & \textbf{MT\#2} & \textbf{MT\#3} & \textbf{MT\#4} & \textbf{MT\#5} \\
    \midrule
    \textbf{EN $\rightarrow$ DE} & & & & & &\\
    WMT 2020(EN $\rightarrow$ DE) & 41.14M & P & P & P & P & P/FT & FT  \\
    OpenSubtitles 2018 & 13.84M & P & P & P & P & P/BT/FT & FT/BT  \\
    MuST-C & 0.23M & P & P/BT & P/BT & P/BT & P/BT/FT & FT/BT \\
    monolingual(EN/DE) &  10.25M & P & BT & BT & BT & BT & BT \\
    \midrule
    \textbf{EN $\rightarrow$ JA} & & & & & &\\
    WMT 2020(EN $\rightarrow$ JA) & 18.19M & P & P/BT & P/BT & P/BT & P/BT/FT & BT/FT  \\
    \bottomrule
  \end{tabular}
  \caption{The statistics of MT datasets and the combination modes of datasets for simultaneous NMT models. ``P" indicates the parallel corpus. ``BT" means the back-translated data generated by MT\#0. ``FT" is the forward-translated data generated by MT\#1-3. }
  \label{table:simul:data2}
\end{table*}

\begin{table}[!t]\small
  \centering
  \begin{tabular}{lcccc}
    \toprule
    \textbf{\#} & \textbf{Model Arch} & \textbf{Enc} & \textbf{Dec} & \textbf{Emb}\\
    \midrule
    0 & Transformer & 6 & 6 & 1024 \\
    1 & Transformer & 6 & 6 & 1024 \\
    2 & Transformer & 50 & 6 & 1024 \\
    3 & LightConv & 6 & 6 & 1024 \\
    4 & Transformer & 16 & 3 & 768 \\
    5 & Transformer & 16 & 3 & 768 \\
    \bottomrule
  \end{tabular}
  \caption{The model setups. ``Enc", ``Dec" denote the number of encoder and decoder layers. ``Emb" means the embedding size and the hidden size.}
  \label{table:simul:model_arch}
\end{table}

\paragraph{Data Augmentation}
Similar to section~\ref{sec:cas_nmt}, we utilize tagged back-translation (BT) and knowledge distillation (KD) strategies to improve the performance of simultaneous NMT.  We experiment with both LightConv~\cite{wu2018pay} and transformer models. The model with the best BLEU score on the development set is chosen for data augmentation. The statistics of all training data and model settings are presented in Table~\ref{table:simul:data2} and Table~\ref{table:simul:model_arch} respectively.

\subsection{Efficient \texttt{wait-k} Model}
\label{simul:model}
% wait-k based; uni-encoder; incremental encoding; multi-path; 
Our simultaneous NMT systems are based on transformer \texttt{wait-k} models, which first read $k$ source tokens and then alternate between reading and writing (translating).
Formally, when decoding the sentence $\mathbf{x}$, the number of visible source tokens is constrained within $\min(k+t-1,|\mathbf{x}|)$ at decoding step $t$, where $k$ is the hyper-parameter controlling the latency.
Furthermore, to avoid recomputing the hidden states of the encoder each time a token is read, we implement incremental unidirectional encoders~\cite{elbayad2020efficient}. And \texttt{multi-path} training is also applied to leverage more possible \texttt{wait-k} paths which refers that hyper-parameter $k \in [3, 9]$ is random selected at each batch during training.

Models are trained with a batch size of 32,000 tokens on Tesla V100 GPUs. We average the last 6 checkpoints once the model converges.

\subsection{Inference}
\label{simul:infer}
We explore the look-ahead beam search strategy for inference. 
Specifically, we apply beam search to generate $M (M > 1)$ tokens at each decoding step and pick the first token in the one with the highest log-probability out of multiple decoding paths. The look-ahead beam search achieves consistent performance improvement when $k_\text{eval}$ is small while its performance improvement is insignificant with a large $k_\text{eval}$. This search method is excluded from our final submissions due to its higher latency, and we choose the greedy search instead.

Additionally, we split the source sentences into sub-sentences once the end-of-sentence punctuation is recognized. Though it may result in a slight performance drop due to the lack of context, we can obtain a much lower latency.

For the final submissions, we use ensemble models.
We train several models with different $k_\text{train}$ values and disjoint subsets of training data for data diversity. Each model produces different latency-quality trade-offs.

\section{Experimental Results}
\label{sec:exp}

% In this section, we report results for offline speech translation with both cascade systems and end-to-end systems,  as well as simultaneous translation.
We conduct all our experiments using NeurST \cite{zhao2020neurst} and report results for the submitted speech translation tasks in this section.
It is worth noting that all transcripts and translations in the test sets are removed from the training data.

When evaluating the offline ST models, tags such as applause and laughing are removed from both hypothesis and reference. 
We use word error rate~(WER) to evaluate the ASR model and report case-sensitive detokenized BLEU\footnote{\url{https://github.com/jniehues-kit/sacrebleu}} for MT. No other data segmentation techniques are applied to the dev/test sets.
Results on MuST-C \textit{dev} and \textit{tst-COMMON}, as well as \textit{dev(v1)} and \textit{tst-COMMON(v1)} from MuST-C v1~\cite{gangi2019} are listed together, which serve as strong baselines for comparison purpose in the end-to-end speech translation field.

\begin{table*}[!t]\small%\setlength{\tabcolsep}{4pt}
\centering
\begin{tabular}{llccccl}
\toprule
\textbf{\#} & \textbf{System} & \textbf{\textit{dev}} & \textbf{\textit{tst-COM}} & \textbf{\textit{dev(v1)}} & \textbf{\textit{tst-COM(v1)}}  & \textbf{Training data composition} \\
\midrule
\multicolumn{2}{l}{\textbf{Pure MT}} & & & & & \\
1 & MT (w/o punc. \& lc) & 32.0 & 34.1 & 32.2 & 34.0 & \multirow{4}{*}{MT~(see Table~\ref{tb:mt_training_sets})} \\
2  & MT (w/ punc. \& tc) & 33.8 & 36.2 & 33.7 & 35.9 & \\
3 & ensemble MT (w/o punc. \& lc) & 33.8 & 35.2 & 33.8 & 35.3 & \\
4 & ensemble MT (w/ punc. \& tc) &  34.7 & 36.7 & 34.6 & 36.2 & \\
\midrule
\multicolumn{2}{l}{\textbf{Cascaded ASR $\rightarrow$ MT}} & & & &  \\
5 & AppTek/RWTH ~\cite{Bahar2020iwslt} & - & - & - & 29.7 & / \\ 
6 & ASR $\rightarrow$ MT & 29.9 & 32.1 & 28.4 & 31.3 & ASR+MT \\
7 & ASR $\rightarrow$ ensemble MT & \textbf{31.7} & \textbf{33.3} & \textbf{30.1} & \textbf{32.3} & / \\
\midrule
\multicolumn{2}{l}{\textbf{End-to-End ST}} & & & &  \\
8 & direct ST baseline & 23.9 & 23.9  & - & - & MuST-C ONLY \\
9 & direct ST & 28.9 & 29.9 & 27.9 & 29.5 & ST+ST Augm. by MT\#1\&2 \\
10 & direct ST++ & 29.6 & \textbf{30.4} & \textbf{28.3} & \textbf{29.7} & ST All \\
11 & direct ST++* & \textbf{30.0} & 30.2 & 28.2 & 29.6 & ST All \\
\midrule
12 & XSTNet-768~\cite{ye2021progressive} & 30.4 & \textbf{31.1} & - & \textbf{30.3} & ASR+MT+ST All \\
13 & direct ST + fbank2vec-512 & 28.7 & 29.1 & 26.7 & 27.6 & ST All \\
14 & PMTL-ST + fbank2vec-768 & 29.6 & 29.6 & 26.9 & 28.1 & ASR+MT+ST All \\
15 & PMTL-ST + fbank2vec-768 ++ & 30.8 & \textbf{31.1} & \textbf{28.8} & 30.1 & ASR+MT+ST All+speed pertub \\
16 & PMTL-ST + fbank2vec-768 ++* & \textbf{30.9} & \textbf{31.1} & \textbf{28.8} & 30.1 & ASR+MT+ST All+speed pertub \\
\midrule
17 & ensemble (9, 10, 11) & 30.4& 31.2 & 29.0 & 30.6 & / \\
18 & ensemble (15, 16) & 31.0 & 31.1 & 28.8 & 30.1 & / \\
19 & ensemble (14, 15, 16) & 31.4 & 31.5 & 29.3 & 30.6 & / \\
20 & ensemble (13, 14, 15, 16) & \textbf{31.6} & \textbf{31.8} & \textbf{29.5} & \textbf{30.8} & / \\
\bottomrule
\end{tabular}
\caption{The overall results of the offline speech translation. The MT model used in the cascade approach is MT\#2 and the ensemble MT model is formed by MT\#2-MT\#5. The direct ST++* is the same as direct ST++ with different random seed for in-domain data over-sampling. The PMTL-ST + fbank2vec-768 ++* is continuously trained from PMTL-ST + fbank2vec-768 ++. \textit{tst-COM} is the abbreviation for \textit{tst-COMMON}.}
\label{tb:offline_results}
\end{table*}

When evaluating the simultaneous translation, we use the official SimulEval \cite{ma2020simuleval} toolkit and report case-sensitive detokenized BLEU \cite{post-2018-call} and Average Lagging \cite{Ma2019stacl} on MuST-C \textit{tst-COMMON} (EN2DE) and IWSLT21 dev set (EN2JA).

\subsection{Offline Speech Translation}
The overall performance of the offline ST and the ASR component used in the cascade system are listed in Table~\ref{tb:offline_results} and Table~\ref{tb:cas_asr_results} respectively.

\begin{table}[!t]\small%\setlength{\tabcolsep}{4pt}
\centering
\begin{tabular}{lc}
\toprule
\textbf{Testset} & \textbf{WER} \\
\midrule
\textit{dev} & 5.2 \\
\textit{tst-COMMON} & 5.7 \\
\textit{dev(v1)} & 10.6 \\
\textit{tst-COMMON(v1)} & 7.4 \\
\bottomrule
\end{tabular}
\caption{The WER of the ASR system for the offline ST.}
\label{tb:cas_asr_results}
\end{table}

In Table~\ref{tb:offline_results}, line 1-4 show the performance of our pure MT systems, which translate the lowercased ground truth transcripts with no punctuation marks, and the natural texts. 
As seen, there may be no essential improvements with the ``$<$ASR$>$" tag on the irregular input (up to 2 BLEU gap on the single model), and it suggests that text restoration has the potential to narrow the gap.
Line 6-7 present the results of translating the ASR output, and we see our cascaded approach surpasses last year's best cascade system (line 5) by 2.6 BLEU. However, there is still a significant loss of up to 3 BLEU scores than line 1/3 due to ASR errors. 

The results of our end-to-end solutions are presented in line 8-20, where line 8 is a benchmark model \cite{zhao2020neurst} trained on the MuST-C dataset only. 
With the growth of model capacity~(256d$\rightarrow$768d) and data augmentation, we obtain 6 BLEU improvement on the \textit{tst-COMMON} over the benchmark (line 8).
Then, increasing the size of augmented data gains slight improvement, as comparing line 9 to line 10/11 (+0.3$\sim$0.5 BLEU scores). Line 13-16 show the results of our proposed fbank2vec.
As shown in line 15, we achieve 31.1 BLEU on \textit{tst-COMMON}, the best single model with fbank2vec, progressive multi-task learning, and speed perturbation. We obtain 31.8 BLEU (line 20) for the final ensemble model, which surpasses the end-to-end benchmark by 7.9 BLEU scores and is approaching the cascade system with a nearly 1.5 BLEU gap.

Lastly, our primary cascade system is line 7, and the primary end-to-end system is line 20 for submission, which achieves higher performance via model ensemble.

\begin{figure*}[tbp]
\centering
\begin{subfigure}[b]{0.45\textwidth}               
\centering                                                \includegraphics[width=\textwidth]{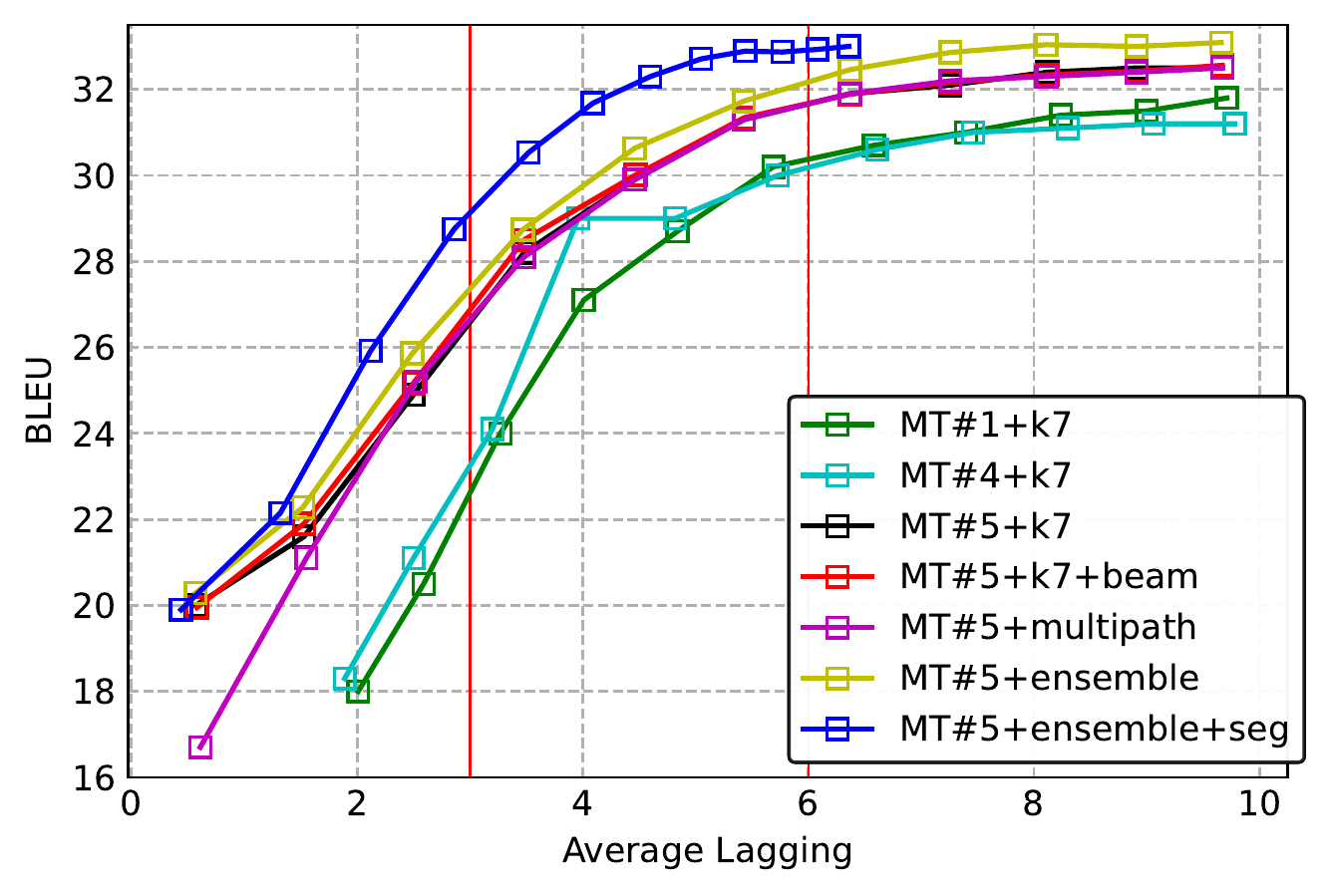}
\caption{EN$\rightarrow$DE}
\label{fig:simul:result:a}
\end{subfigure}
\begin{subfigure}[b]{0.45\textwidth}   
%第二张子图
\centering                      
\includegraphics[width=\textwidth]{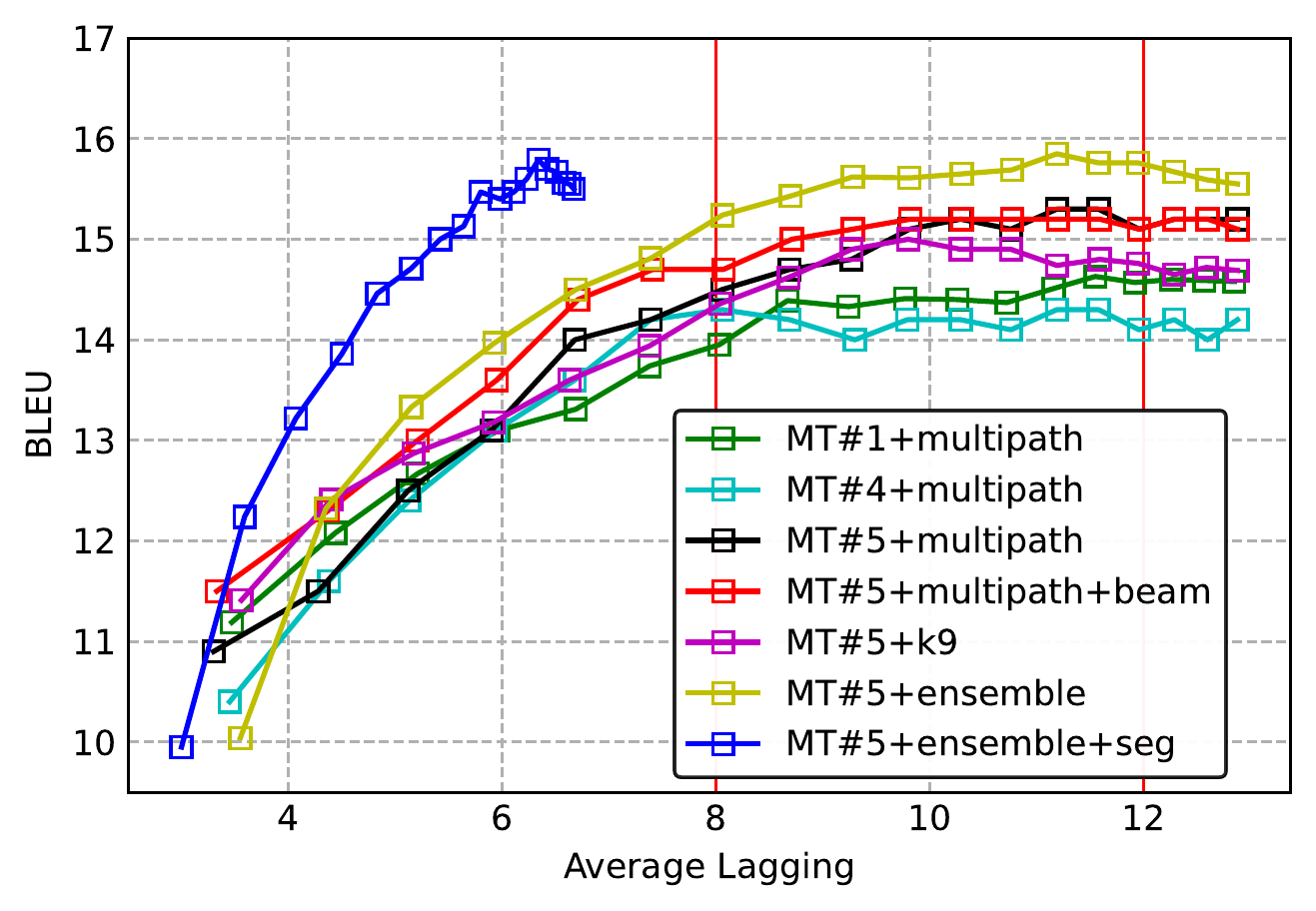}
\caption{EN$\rightarrow$JA}
\label{fig:simul:result:b}
\end{subfigure}
\caption
{Latency-quality trade-offs of the simultaneous NMT. $k7/9$ means $ {k}_\text{train}=7/9$. MT\#X indicate the aforementioned training datasets and model settings in Table~\ref{table:simul:data2} and \ref{table:simul:model_arch}. \texttt{beam} refers to our look-ahead beam search strategy. \texttt{seg} means that the sentences are pre-splited during inference. \texttt{multipath} means that $k$ is random selected during training.}
\label{fig:simul:result}
\end{figure*}

\begin{table}[t]\small
	\centering
	\begin{tabular}{ccccc}
		\toprule
		 & & Low & Medium & High \\  
		\midrule
		\multirow{2}{*}{EN $\rightarrow$ DE} & Ensemble & 25.86 & 31.73 & 33.21 \\
		& +seg & 28.75 & 32.87 & 32.97 \\
	    \midrule
		\multirow{2}{*}{EN $\rightarrow$ JA} & Ensemble & 14.81 & 15.85 & 15.85 \\
		& +seg & 15.79 & 15.79 & 15.79 \\
		\bottomrule
	\end{tabular}
	\caption{Performance of our final submissions models on MuST-C \textit{tst-COMMON} for English-German and IWSLT21 dev set for English-Japanese.}
	\label{table:simul:result}
\end{table}

\subsection{Simultaneous Translation}
\label{sec:exps:simul}
% plot; different k ; different data ; ensemble ; force-segment
We evaluate the simultaneous NMT systems with different combinations of strategies and present our results in Figure~\ref{fig:simul:result}.
Then we report the performance on different latency regimes in Table \ref{table:simul:result}.

As shown in Figure \ref{fig:simul:result}, we can obtain remarkable BLEU improvements by training with only the knowledge distilled data (black) comparing to the filtered parallel data (green) and back-translated data (magenta), on average 1.0 BLEU improvement on EN$\rightarrow$DE and 0.5 on EN$\rightarrow$JA.
The possible reasons may be: 1) Noise in origin data is migrated, like non-parallel sentence pairs. 2) Complex sentences with diverging word order are excluded, and the machine-translated texts, i.e., translationese, sometimes have simpler expressions. 

We can see that the proposed look-ahead beam search (red) is competitive when $k_\text{eval}$ is relatively small but is comparable with the greedy search when $k_\text{eval}$ is large. So overall considering translation latency, we use the greedy search for our final submissions. As for \texttt{multi-path} training, we see it achieves limited BLEU improvement in our experiments. 

For our final submission of EN$\rightarrow$DE, we use the ensemble model, which consists of three transformer models trained on different dataset combinations, with $k_\text{train}=7$. For EN$\rightarrow$JA, the submitted model is formed by two transformer models, with $k_\text{train}=\infty$ (trained on full sentences) and \texttt{multi-path} training respectively. As presented in Figure~\ref{fig:simul:result}, the model ensemble technique leads to at least 0.5 BLEU improvement on average (yellow). Additionally, with the sentence segmentation (bleu), the average lagging is significantly reduced. 
As a result, our final submitted systems exceed the baseline system at around 7 BLEU on the same latency regime.

\section{Final Results}
\label{sec:final}

\begin{table}[t]\small
	\centering
	\begin{tabular}{lcccc}
		\toprule
		\multirow{2}{*}{\# - System} & \multirow{2}{*}{tst2020} & \multicolumn{3}{c}{tst2021} \\  
		 & & ref2 & ref1 & both\\
		\midrule
		\textbf{7 - Cascade (ensemble)} & 22.2 & 21.8 & 17.1 & 29.5 \\
		6 - Cascade (single) & 21.0 & 20.3 & 16.4 & 27.7 \\
		\midrule
		\textbf{20 - Direct (ensemble)} & 24.3 & 21.7 & 18.7 & 31.3 \\
		16 - Direct (single) & 23.5 & 21.6 & 18.2 & 30.6 \\
		17 - Direct (ensemble) & 22.4 & 21.1 & 17.5 & 29.2 \\
		10 - Direct (single) & 21.6 & 20.4 & 17.0 & 28.1 \\
		\bottomrule
	\end{tabular}
	\caption{BLEU of the IWSLT 2021 submissions for offline speech translation task. The rows in bold are our primary systems. The \texttt{ref1} of tst2021 is originally from the TED website, while the \texttt{ref2} is newly created for this year's campaign.}
	\label{table:offline:blind_result}
\end{table}

\begin{table}[t]\small
	\centering
	\begin{tabular}{lccccc}
		\toprule
		System & BLEU & AL & AP & DAL \\  
		\midrule
		\textbf{EN $\rightarrow$ DE} & & & & & \\
		MT(Low Latency) & 23.24 & 3.08 & 0.68 & 4.25 \\
        MT(Mid Latency) & 27.22 & 6.30 & 0.81 & 9.24 \\
        MT(High Latency) & 26.82 & 12.03 & 0.92 & 12.39 \\
	    \midrule
	    \textbf{EN $\rightarrow$ JA} & & & & & \\
	    MT(Low Latency) & 16.91 & 6.54 & 0.89 & 11.26 \\
        MT(Mid Latency) & 16.91 & 6.54 & 0.89 & 11.26 \\
        MT(High Latency) & 16.97 & 11.27 & 0.97 & 11.90 \\
		\bottomrule
	\end{tabular}
	\caption{Performance of the IWSLT 2021 submissions for simultaneous NMT on the blind test set.}
	\label{table:simul:blind_result}
\end{table}

Table~\ref{table:offline:blind_result} lists the final results of the IWSLT 2021 offline ST track. Surprisingly, we find that our end-to-end models significantly surpass the cascade systems, which is different from our conclusions on the MuST-C test sets. We think this may be caused by the reference of tst2021. Since the \texttt{ref1} of tst2021 is the original one from the TED website, the translations could be much shorter for subtitling, and our end-to-end models may fit well on it.

Table~\ref{table:simul:blind_result} shows the official evaluation for our simultaneous NMT systems. 

\section{Conclusion}
\label{sec:conclusion}
This paper summarizes the results of the shared tasks in the IWSLT 2021 produced by the Volctrans team. We investigate the performance of the end-to-end solutions with data augmentation and progressively training framework for the offline ST task. Our end-to-end approach surpasses the last year's best cascaded system by 1 BLEU, but it is still lagging behind our cascade model by 1.5 BLEU scores on MuST-C test sets. However, our end-to-end solutions achieve promising performance on tst2020 and tst2021.
Afterwards, we develop the efficient \texttt{wait-k} model with \texttt{multi-path} training, and large-scale knowledge distillation and back translation methods. The final submitted systems exceed the baseline systems at 7 BLEU on the same regime. We see the data augmentation technique plays the most important role in these tasks.
In the future, we would like to explore a more extensive data condition on both modality and quantity.
We hope our practice could facilitate batch research works and industrial applications.

% \section*{Acknowledgments}

% The acknowledgments should go immediately before the references. Do not number the acknowledgments section.
% \textbf{Do not include this section when submitting your paper for review.}

\bibliographystyle{acl_natbib}
% \bibliography{anthology,acl2021}
\bibliography{acl2021}
%\appendix

\end{document}